\documentclass{article}

\usepackage{microtype}
\usepackage{graphicx}
\usepackage{subfigure}
\usepackage{booktabs} 

\usepackage{hyperref}

\usepackage{algorithmic}
\usepackage{enumitem}
\renewcommand{\algorithmicrequire}{\textbf{Input:}} 
\renewcommand{\algorithmicensure}{\textbf{Output:}} 

\usepackage[accepted]{icml2024}

\usepackage{amsmath}
\usepackage{amssymb}
\usepackage{mathtools}
\usepackage{amsthm}

\usepackage[capitalize,noabbrev]{cleveref}

\theoremstyle{plain}
\newtheorem{theorem}{Theorem}[section]

\theoremstyle{definition}

\theoremstyle{remark}

\usepackage[textsize=tiny]{todonotes}

\renewcommand{\epsilon}{\varepsilon}

\newcommand{\loss}{\mathcal{L}}
\renewcommand{\L}{\mathcal{L}}
\newcommand{\mseloss}{\textrm{MSE loss}}
\newcommand{\celoss}{\textrm{CE loss}}
\newcommand{\target}{\textrm{target}}

\newcommand{\argmin}[1]{\textrm{argmin}\left\{#1\right\}}
\newcommand{\textblue}[1]{\textcolor{blue}{#1}}

\newcommand{\bra}{\langle}
\newcommand{\ket}{\rangle}
\newcommand{\dom}{\mathrm{dom}}
\newcommand{\setB}{\mathcal{B}}

\newcommand{\setE}{\mathcal{E}}
\newcommand{\R}{\mathbb{R}}

\def\ba{a}

\def\bc{c}
\def\bu{u}
\def\bv{v}

\icmltitlerunning{Understand the Effectiveness of Shortcuts through the Lens of DCA}

\begin{document}

\twocolumn[
\icmltitle{Understand the Effectiveness of Shortcuts through the Lens of DCA}



\icmlsetsymbol{equal}{*}

\begin{icmlauthorlist}
\icmlauthor{Youran Sun}{equal,ymsc}
\icmlauthor{Yihua Liu}{equal,ymsc}
\icmlauthor{Yi-Shuai Niu}{equal,bimsa}
\end{icmlauthorlist}

\icmlaffiliation{ymsc}{Yau Mathematical Sciences Center (YMSC), Tsinghua University, Beijing, China}
\icmlaffiliation{bimsa}{Beijing Institute of Mathematical Sciences and Applications (BIMSA), Beijing, China}

\icmlcorrespondingauthor{Yi-Shuai Niu}{niuyishuai@bimsa.cn}

\icmlkeywords{Machine Learning, ICML, Optimization, Neural Network Architecture}

\vskip 0.3in
]



\printAffiliationsAndNotice{\icmlEqualContribution} 

\begin{abstract}
Difference-of-Convex Algorithm (DCA) is a well-known nonconvex optimization algorithm for minimizing a nonconvex function that can be expressed as the difference of two convex ones.
Many famous existing optimization algorithms, such as SGD and proximal point methods, can be viewed as special DCAs with specific DC decompositions, making it a powerful framework for optimization.
On the other hand, shortcuts are a key architectural feature in modern deep neural networks, facilitating both training and optimization. We showed that the shortcut neural network gradient can be obtained by applying DCA to vanilla neural networks, networks without shortcut connections.
Therefore, from the perspective of DCA, we can better understand the effectiveness of networks with shortcuts.
Moreover, we proposed a new architecture called NegNet that does not fit the previous interpretation but performs on par with ResNet and can be included in the DCA framework.
\end{abstract}

\section{Introduction}\label{introduction}

The difference of Convex Algorithm (DCA) was first introduced by Pham Dinh Tao in 1985 as an extension of the subgradient algorithm and then extensively developed by Pham Dinh Tao and Le Thi Hoai An in the 1990s for efficiently solving both smooth and nonsmooth nonconvex global optimization problems \cite{dca1997,dca1998,dca2016,sdca2022,sdca2022-2}.
DCA is concerned with the following standard DC optimization problem
\begin{equation*}
    \min\{F(x):=G(x)-H(x)\mid x\in \R^n\}
\end{equation*}
in which $G$ and $H$ are proper and closed convex functions.
Almost all functions encountered in optimization can be expressed as difference-of-convex (DC).
A DC decomposition of $\cos(x)$ can be expressed as:
\begin{equation*}  
    \cos(x) = \left(\cos(x) + \frac{1}{2}x^2\right) - \frac{1}{2}x^2.  
\end{equation*}  
Moreover, DC decompositions based on sums-of-squares (referred to as the DC-SOS decomposition) for general polynomials have been established in \cite{niu2024difference,niu2024power}.
Over the past 40 years, the DCA has been successfully applied to a wide range of nonconvex optimization problems, including trust-region subproblems, nonconvex quadratic programs, and various real-world applications spanning areas such as molecular conformation \cite{molecular2003}, network optimization \cite{network2002}, portfolio optimization \cite{pham2011efficient}, eigenvalue optimization \cite{niu2019improved}, sentence compression \cite{niu2021difference}, and combinatorial optimization \cite{comb2001}, among others.

On the other hand, people are training deeper and deeper neural networks.
Having more layers makes it easier to learn complex and abstract features.
However, deep neural networks are notoriously hard to train.
Sometimes, the generalization ability of a deeper network is even worse than that of a shallower network.
Shortcut, or residual, is one of the few techniques to cure this problem
and is now a standard component in deep neural networks.
\cite{resnet2015} was the first paper to demonstrate the power of shortcuts. 
They introduced ResNet, which is characterized by incorporating many shortcuts into CNNs for image classification tasks and achieved unprecedented accuracy. 
Currently, all mainstream neural network architectures, including ResNet \cite{resnet2015}, UNet \cite{unet}, and Transformer \cite{transformer}, use shortcuts to help with training.

In this work, we analyze neural networks from the perspective of DCA.
We find a DC decomposition of the vanilla networks (networks without
shortcut connections), which can give rise to shortcut network gradients for both $\mseloss$ and $\celoss$.
That is to say
\textbf{
$$    \textrm{DCA}+\textrm{vanilla network}=\textrm{SGD}+\textrm{shortcut network}.$$}
In other words, one can invent ResNet by applying DCA to vanilla CNN without knowing the shortcuts.
Previous explanations of shortcuts focus on the information flow (the zero-order derivative) or the gradient (the first-order derivative).
However, this paper introduces a novel insight, positing that shortcuts' true strength lies in their implicit utilization of second-order derivative information, even within the confines of first-order optimization methods.
This refined understanding prompts a reevaluation of architectural designs in neural networks, emphasizing the integration of second-order information to enhance convexity and, consequently, the overall performance of the learning algorithm.
We propose that one can design new architectures by applying DCA to existing architectures, and the DCA theorems will be useful in proving the global convergences of the neural network training process via SGD.

The organization of the paper is as follows.
Section \ref{sec:dca} reviews the basics of DCA and gives some useful examples.
We then proceed to the main results in Section \ref{sec:main}.
In Section \ref{sec:negnet}, we demonstrate the usage of DCA by a simple experiment called ``NegNet" (Negative ResNet).
In Section \ref{sec:relatedwork}, we review the theoretical works that explain the effectiveness of shortcuts.
Finally, we provide a conclusion and outlook in Section \ref{sec:conclusion}.

\section{Introduction to DCA}
\label{sec:dca}

DCA is a well-known optimization approach.
Many existing methods can be just viewed as a special DCA.
It is a very powerful philosophy to generate algorithms. 
Its object is to optimize
\begin{equation}\label{prob:P}
    \min \{F(x):=G(x)-H(x) \mid x\in \R^n\}, \tag{P}
\end{equation}
where $G$ and $H$ are both convex.
The main idea of DCA is simple: at each step $k$, linearizing the 2nd DC component $H(x)$ at $x^k$ 
\begin{equation}\label{eq:affineH}
    H_k(x):=H(x^k)+\bra x-x^k,y^k\ket,
\end{equation}
and minimizing the resulting convex function
$$F_k(x):=G(x) - H_k(x).$$
The $y^k$ in Eq. \eqref{eq:affineH} should be chosen in the subdifferential
\begin{equation*}
    \partial H(x^k):=\{y \mid H(x) \geq H(x^k) + \bra y, x-x^k \ket, \forall x\}.
\end{equation*}
The DCA iteration is
\begin{equation}\label{eq:dcaiter}
\begin{aligned}
    &\begin{cases}
    y^{k \phantom{+1}}\in \partial H(x^k) \\
    x^{k+1} \in \argmin{F_k(x)} = \argmin{G(x)-\bra x,y^k\ket}
    \end{cases}\\
\end{aligned}
\end{equation}

\begin{algorithm}[htb!]
   \caption{DCA}
   \label{alg:dc}
\begin{algorithmic}
    \REQUIRE Initial point $x^0$, DC decomposition $G$ and $H$.
   \FOR{$k=0,1,\ldots$}
   \STATE Calculate $y^k\in \partial H(x^k)$.
   \STATE Calculate $x^{k+1}\in \argmin{G(x)- \bra x,y^k\ket}$.
   \ENDFOR
\end{algorithmic}
\end{algorithm}

DCA is summarized in Algorithm \ref{alg:dc}.
The convergence of DCA is well guaranteed.
\begin{theorem}[Convergence Theorem of DCA, see e.g., \cite{dca1997,niu2022convergence}]\label{thm:convDCA}
Let \(\{x^k\}\) and \(\{y^k\}\) be the sequences generated by DCA for the DC problem \eqref{prob:P}, starting from an initial point \(x^0 \in \dom~\partial H\). Suppose that both \(\{x^k\}\) and \(\{y^k\}\) are bounded. Then:
\begin{itemize}[itemsep=2pt, topsep=2pt]
    \item[$\bullet$] The sequence \(\{F(x^k)\}\) is non-increasing and bounded from below, and thus convergent to some limit \(F^*\).
    \item[$\bullet$] Every cluster point \(x^*\) of the sequence \(\{x^k\}\) is a \emph{DC critical point}, i.e., \(\partial G (x^*) \cap \partial H(x^*) \neq \emptyset\).
    \item[$\bullet$] If \(H\) is continuously differentiable on \(\mathbb{R}^n\), then every cluster point \(x^*\) of the sequence \(\{x^k\}\) is a \emph{strongly DC critical point}, i.e., \(\nabla H(x^*) \in \partial G (x^*)\).
\end{itemize}
\end{theorem}

We will end this section with an example of DCA.

\paragraph{Example: }
For a differentiable target function $F(x)$ such that $F(x) + \frac{\rho}{2}\|x\|^2$ is convex for some large enough $\rho$, we have the \textbf{Proximal DC Decomposition}
\begin{equation*}
    G_1(x)=F(x)+\frac{\rho}{2}\|x\|^2,\; H_1(x)=\frac{\rho}{2}\|x\|^2.
\end{equation*}
The DCA for the above decomposition is
\begin{equation*}\begin{aligned}
    y^k&=\rho x^k,\\
    x^{k+1}&\in\argmin{F(x)+\frac{1}{2}\rho\|x\|^2-\bra x,\rho x^k \ket}\\
    &=\left(I+\frac{1}{\rho}\nabla F\right)^{-1}(x^k)
\end{aligned}\end{equation*}
This is nothing else but the Proximal Point Algorithm (PPA).

Another well-known DC decomposition for a differentiable target function $F(x)$ such that $\frac{\rho}{2}\|x\|^2 - F(x)$ is convex for some large enough $\rho>0$ is called the \textbf{Projective DC Decomposition} given by
\begin{equation*}
    G_2(x)=\frac{\rho}{2}\|x\|^2,\; H_2(x)=\frac{\rho}{2}\|x\|^2 -F(x).
\end{equation*}
The corresponding DCA is 
\begin{equation*}\begin{aligned}
    y^k&=\rho x^k - \nabla F(x^k),\\
    x^{k+1}& = \argmin{\frac{\rho}{2}\|x\|^2 - \bra x,\rho x^k - \nabla F(x^k) \ket}\\
    &=x^k -\frac{1}{\rho}\nabla F(x^k)
\end{aligned}\end{equation*}
This is the Gradient Descent algorithm (GD) with a learning rate $\textrm{lr}=1/\rho$. 
Note that the DC structure provides an exciting insight for choosing the learning rate: $\textrm{lr}$ should ensure $\frac{1}{2\,\textrm{lr}}\|x\|^2-F(x)$ convex.

This example shows the unique ability of DCA to unify existing convex optimization algorithms.
Many methods and optimizers can be seen as DCA under particular DC decomposition, and their convergences were proved all-in-one in the DCA convergence theorem.

As a concrete example, consider the following nonconvex quadratic optimization problem
\begin{equation*}
    \min\left\{\left.\frac{1}{2}x^{\top} A x+b^{\top} x\;\right|
    x\in\setB
    \right\},
\end{equation*}
where $A$ is symmetric but not positive semi-definite and $\setB$ is the Euclidean ball defined by $\{x\mid \|x\|\leq r\}$.
We can select the proximal DC decomposition as follows
\begin{equation*}
    G_1(x)=\frac{1}{2}x^{\top}(A+\rho I)x+b^{\top} x, H_1(x)=\frac{1}{2}\rho x^{\top} x,
\end{equation*}
resulting in the update
\begin{equation*}
    x^{k+1}=\argmin{\frac{1}{2}x^\top (A+\rho I) x+\bra b-\rho x^k,x\ket \mid x\in \setB }.
\end{equation*}
By letting $z = (A+\rho I)^{\frac{1}{2}} x$, this convex quadratic problem is equivalent to
\begin{equation*}\begin{aligned}
    &\argmin{\frac{1}{2}\|z\|^2+\bra b-\rho x^k,(A+\rho I)^{-1/2}z\ket \mid z\in\setE}\\
    &=P_{\setE}\left((A+\rho I)^{-1/2}(\rho x^k - b)\right);
\end{aligned}\end{equation*}
thus
\begin{equation*}
    x^{k+1}=(A+\rho I)^{-1/2}P_{\setE}\left((A+\rho I)^{-1/2}(\rho x^k - b)\right)
\end{equation*}
where $\setE$ is the ellipsoid $\{z\mid z^\top (A+\rho I)^{-1} z\leq r^2\}$ and $P_\setE$ denotes the projection of a vector onto the ellipsoid $\setE$. 

Alternatively, we can choose the Projective DC Decomposition
\begin{equation*}
    G_2(x)=\frac{1}{2}\rho x^{\top} x+b^{\top} x,\quad H_2(x)=\frac{1}{2}x^{\top}(\rho I-A) x.
\end{equation*}
The DCA iteration then becomes
\begin{equation*}
    x^{k+1}=P_\setB\left(x^k-\frac{1}{\rho}(A x^k +b)\right),
\end{equation*}
where $P_\setB$ is the projection onto the Euclidean ball $\setB$.
This algorithm is guaranteed to converge to the global minimum with suitable restarts, as demonstrated in \cite{dca1998}.


\section{From DCA to Shortcuts}
\label{sec:main}

In this section, we discuss how applying DCA to a vanilla network leads to the gradient structure of ResNet.

We denote a layer by \(F^l\), where \(l\) is the layer index, and \(L\) represents the last layer. The inputs to \(F^l\) are \(h^l\) and \(w^l\), where \(w^l\) includes both the weights and biases. Note that \(F^l\) depends implicitly on \(w^m\) for \(m < l\) through \(h^l\).

Let \(\partial_{\text{var}}\) denote the derivative with respect to the variable \(\text{var}\). Using \(\#h\) to represent the hidden layer width and \(\#w\) to denote the number of parameters, the derivative \(\partial_w F^l\) is a tensor with shape \((\#h, \#w)\). Specifically, we have:
\[
\left(\partial_w F^l\right)_{i,\alpha} = \partial_{w^l_\alpha} F^l_{i}, \quad
\partial_{w^{L-1}} \loss = \partial_h \loss \, \partial_w F^{L-1},
\]
where \(\loss\) represents the loss function. One advantage of this notation is that the chain rule applies from left to right. Let \(*X.\text{shape}\) denote the shape for a high-order tensor \(X\). Then the shape of \(\partial_h X\) is \((*X.\text{shape}, \#h)\), and the shape of \(\partial_w X\) is \((*X.\text{shape}, \#w)\).

\begin{figure}[h]
    \centering
    \includegraphics[width=1.0\linewidth]{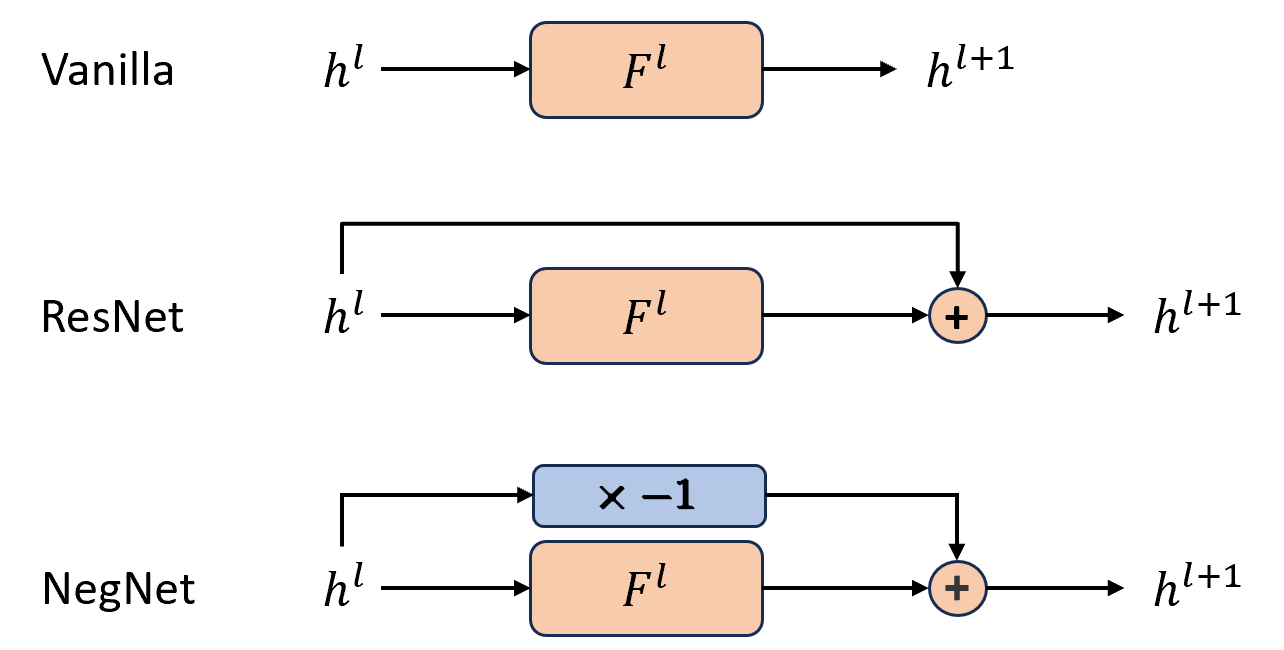}
    \caption{The architectures of the vanilla network, ResNet, and the NegNet proposed in this paper.
    }
    \label{fig:nets}
\end{figure}

The vanilla neural network paradigm is
\begin{equation}\begin{cases}
    h^1=F^0(x^0,w^0),~\cdots,\\
    h^l=F^{l-1}(h^{l-1},w^{l-1}),~\cdots,\\
    h^L=F^{L-1}(h^{L-1},w^{L-1}),\\
    \loss=\loss(h^L,\target),
\end{cases}\end{equation}
where $x_0$ is the input.
While the new paradigm after ResNet is
\begin{equation}\label{eq:resnet}\begin{cases}
    h^1=x^0+F(x^0,w^0)=:x^0+F^0(x^0),~\cdots\\
    h^l=h^{l-1}+F(h^{l-1},w^{l-1})=:h^{l-1}+F^{l-1}(h^{l-1}),~\cdots\\
    h^L=h^{L-1}+F^{L-1}(h^{L-1}),\\
    \loss=\loss(h^L,\target).
\end{cases}
\end{equation}
It is characterized by (a) shortcuts and (b) every layer having the same structure, differing only in parameters.
Usually, the $F$ in ResNet is not a single layer but $2\sim4$ layers with activation functions.
The gradient of the vanilla network is
\begin{equation}\begin{aligned}
    \partial_{w^{L-1}}\L&=\partial_{h}\L\,\partial_w F^{L-1}\\
    \partial_{w^{L-2}}\L&=\partial_{h}\L\,\partial_{h}F^{L-1}\,\partial_w F^{L-2}\\
    \partial_{w^{L-n}}\L&=\partial_{h}\L\,\prod_{m=1}^{n-1}\partial_{h}F^{L-m}\,\partial_w F^{L-n}.
\end{aligned}
\end{equation}
While the gradient for ResNet is
\begin{equation}\label{eq:resnetgrad}\begin{aligned}
    \partial_{w^{L-1}}\L&=\partial_{h}\L\,\partial_w F^{L-1}\\
    \partial_{w^{L-2}}\L&=\partial_{h}\L\,(I+\partial_{h}F^{L-1})\,\partial_w F^{L-2}\\
    \partial_{w^{L-n}}\L&=\partial_{h}\L\,\prod_{m=1}^{n-1}(I+\partial_{h}F^{L-m})\,\partial_w F^{L-n}.
\end{aligned}
\end{equation}

We want to apply DCA on the vanilla network and get the gradient of ResNet.
For that purpose, we need to analyze the 2nd-order derivative of the vanilla network
{\small
\begin{equation*}\begin{aligned}
    \partial_{w^{L-1}}^2 \L &= 
    \left(\partial_{w}F^{L-1}\right)^{\top}
    \textblue{\partial_{h}^2 \L}\,
    \partial_{w}F^{L-1},\\
    \partial_{w^{L-2}}^2 \L &= 
    \left(\partial_{h}F^{L-1}\partial_{w}F^{L-2}\right)^{\top}
    \textblue{\partial_{h}^2 \L}\\
    &\hspace{7em}\left(\partial_{h}F^{L-1}\partial_{w}F^{L-2}\right),\\
    \partial_{w^{L-1}}\partial_{w^{L-2}} \L &=
    \left(\partial_{h}F^{L-1}\partial_{w}F^{L-2}\right)^{\top}
    \textblue{\partial_{h}^2 \L} \,
    \partial_{w}F^{L-1}\\
    &+\left(\partial_{w}F^{L-2}\right)^{\top} \partial_{h} \L \,\partial_{w}\partial_{h}F^{L-1}.
\end{aligned}
\end{equation*}
}
\noindent Here, we assume that the activation function is ReLU, and we use the property $\partial^2_h F^l = \partial^2_w F^l = 0$ of ReLU.
The above 2nd-order derivative can be further cast into the form of a matrix
\begin{equation}
    \partial^2_{\left(w^{L-1},w^{L-2}\right)}\L=A^{\top}\textblue{\partial_{h}^2 \L} \,A+
    \begin{pmatrix}
        &\ast^{\top} \\ \ast
    \end{pmatrix},
\end{equation}
where
\begin{equation}\begin{aligned}
    A:=&\begin{pmatrix}
        \partial_{w}F^{L-1}&
        \partial_{h}F^{L-1}\partial_{w}F^{L-2}
    \end{pmatrix},\\
    \ast:=&\left(\partial_{w}F^{L-2}\right)^{\top} \partial_{h} \L \,\partial_{w}\partial_{h}F^{L-1}.
\end{aligned}\end{equation}
The first matrix is positive definite because the loss is convex $\partial_{h}^2 \L\succ 0$.
However, the second matrix is not positive definite because the eigenvalues of the matrix $\begin{pmatrix}  & \ast^{\top} \\ \ast &  \end{pmatrix}$ appear in positive and negative pairs.

One can always do the following DC decomposition
\begin{equation}\label{eq:trivialdc}
    G_1=\lambda L_2,\;H_1=\lambda L_2-\L,
\end{equation}
where $L_2$ is the L2 regularization on parameters.
As long as $\lambda$ is large enough\footnote{In machine learning practice, parameters are often constrained within a limited range, so this $\lambda$ always exists.}, $G_1$ and $H_1$ will be both convex, and the Stochastic DCA \cite{sdca2022,sdca2022-2} reduces back to SGD (Stochastic Gradient Descent).

The SGD approximates the low-rank Hessian matrix using a full-rank identity matrix, which we consider to be a rather crude approximation.
Thus, we believe a more accurate estimation is necessary.
To cancel the negative part, consider the following ``counter loss"
\begin{equation}
    \L'_{\alpha}=\L\left(\alpha h^L+(1-\alpha) h^{L-1}\right).
\end{equation}
The Hessian for $\L'_{\alpha}$ reads
$\partial^2_{w^{L-1},w^{L-2}}\L'_{\alpha}=$
\begin{equation*}
    (\alpha A+(1-\alpha) C)^{\top}\partial_{h}^2 \L' \,(\alpha A+(1-\alpha) C)+
    \alpha\begin{pmatrix}
        &\ast^{\top} \\ \ast
    \end{pmatrix},
\end{equation*}
where
\begin{equation}
    C:=\begin{pmatrix}
        0&\partial_{w}F^{L-2}
    \end{pmatrix}.
\end{equation}
It is easy to see that $\L'_{-1}$ can absorb the negative eigenvalues in the $*$ part.
So, we propose the following DC decomposition
\begin{equation}
    G_2=\L+\frac{1}{2}\L'_{-1}+\frac{\rho}{2}L_2,\;H_2=\frac{1}{2}\L'_{-1}+\frac{\rho}{2}L_2.
\end{equation}
The $\rho$ should be larger than the largest negative eigenvalue of the $*$ matrix and is relatively small \footnote{In our practice with CIFAR10, $\rho$ is often less than 5\% of $\lambda$.} compared to the $\lambda$ in Eq. \eqref{eq:trivialdc}. 
What is the DC iteration corresponding to this DC decomposition?
Recall that DC's philosophy is to linearize $H$ at each step, so the convex subproblem is
\begin{equation}
    \argmin{G_2-\textrm{Linearize}(H_2)}.
\end{equation}
Using Newton's method to solve this subproblem is equivalent to
\begin{equation}\label{eq:dcaiterG2}
    \argmin{\partial_w \L\, \Delta w+\frac{1}{2}\Delta w^{\top} \partial_w^2 G_2 \Delta w}
\end{equation}
where $\partial_w \L=\partial_h \L A$ and
\footnote{
We ignore the $*$ term here, which is not an approximation.
In fact, based on our subsequent discussion of high-dimensional vector orthogonality, the eigenvectors corresponding to the $*$ part are perpendicular to the $\ba,\bc$ and therefore, will not affect the final solution.
This is the ``blessing from the high dimension".
},
\begin{equation*}
    \partial_w^2 G_2 = A^{\top}\partial_h^2 \L A +\frac{1}{2}(2C-A)^{\top}\partial_h^2 \L'(2C-A)+\frac{\rho}{2}.
\end{equation*}

\paragraph{MSE Loss Case:}Let's first discuss mean square error (MSE) Loss where $\L(h)=(h^{\top} w^L+b^L-\target)^2=:r^2$.
In this case
\begin{equation}\label{eq:mse1ststep}
    \partial_h \L=2 r w^L,\quad \partial^2_h \L=2 w^L (w^L)^{\top}.
\end{equation}
So apart from the $\rho$ term, $\partial_w^2 G_2$ is rank 2
\begin{equation}\label{eq:mse2ndstep}
    \frac{1}{2}\partial_w^2 G_2-\frac{\rho}{4} = \ba\,\ba^{\top} +\frac{1}{2}(2\bc-\ba)(2\bc-\ba)^{\top},
\end{equation}
with
\begin{equation}
    \ba:=A^{\top} w^L,\quad \bc:=C^{\top} w^L.
\end{equation}
Now, it is time to use the magic property of deep neural networks.
We all know that the effectiveness of contemporary neural networks emerges when they have a super large number of parameters.
Therefore, any theory explaining why neural networks work should incorporate the mentioned characteristics.
However, as far as we know, few works used this fact.
In our work, we regard the large number of parameters as high-dimensional space.
When it comes to very high-dimensional (Euclidian) space, the first thing that comes to our mind is that two random vectors are orthogonal.
In fact, for 2 random unit vectors $\bu$ and $\bv$ in $N$ dimensional space
\begin{equation}
    \mathbb{E}\left[(\bu\cdot\bv)^2\right]=\frac{1}{N}\xrightarrow{N\to\infty} 0.
\end{equation}
Or, more romantically, the surface area of a high-dimensional ball concentrates near its equator.
As a consequence, the volume of a unit ball tends to 0 as the dimension tends to infinity.

So we can view $\ba$ and $\bc$ as orthogonal vectors and the solution to problem \eqref{eq:dcaiterG2} is
\begin{equation}
    \Delta w\propto \ba+\frac{\|a\|^2}{2\|c\|^2+\rho/4}\bc.
\end{equation}
We can see that apart from the original gradient $\ba$, there is a component parallel to the shortcut gradient $\bc$.
This essentially represents the gradient of a ResNet \eqref{eq:resnetgrad}.
Thus, if we begin with a vanilla network and wish to apply DCA, it makes sense to design a network with shortcut connections.
This design choice allows us to compute $\bc$ using automatic differentiation tools.
This highlights the crucial role of the DCA approach in designing new network architectures.

\paragraph{CE Loss Case:}The cross-entropy (CE) Loss case needs a little approximation, but the approximation is the same for SGD and DCA.
The canonical form of CE Loss is
\begin{equation*}
    \L(h)=-\ln{\frac{e^{E_t}}{\sum_{k=1}^{C}e^{E_k}}}=-\left(h^{\top} w^L_{t}+b^L_{t}\right)+\ln{\sum_{k=1}^{C}e^{E_k}},
\end{equation*}
where $E_k=h^{\top} w^L_{k}+b^L_{k}$ is the energy\footnote{Actually, negative energy and the temperature $\beta=1$.} of the $k$-th class, $C$ is the number of classes, $E_t$ is the energy of the target class.
Its derivative is
\begin{equation}
    \partial_{h} \L=-w^L_t+\frac{1}{\sum_{l=1}^C e^{E_l}}\sum_{k=1}^{C}e^{E_k}w^L_{k},
\end{equation}
and its 2nd-order derivative is
\begin{equation}\begin{aligned}
    \partial^2_{h} \L
    &=\frac{1}{\left(\sum_{m=1}^C e^{E_m}\right)^2}
    \sum_{k,l}e^{E_k+E_l}w_{k}(w_{k}-w_{l})^{\top}\\
    &=\frac{1}{\left(\sum_{m=1}^C e^{E_m}\right)^2}
    \sum_{k<l}e^{E_k+E_l}(w_{k}-w_{l})(w_{k}-w_{l})^{\top}.
\end{aligned}\end{equation}
Albert complex, an interesting relation between the 1st-order and 2nd-order derivatives exists
\begin{equation*}
    (\partial_{h} \L+w_t^L)(\partial_{h} \L+w_t^L)^{\top}+\partial_{h}^2 \L
    =\frac{\sum_{k=1}^{C}e^{E_k}w_{k}w_{k}^{\top}}{\sum_{m=1}^C e^{E_m}}.
\end{equation*}
When $E_t\gg E_{i\ne t}$, $\partial_h^2 \L$ is approximately proportional to $\partial_h \L(\partial_h \L)^{\top}$
\begin{equation}
    \partial_h^2 \L \approx \frac{e^{E_t}}{\sum_{i\ne t}e^{E_i}}\partial_h \L(\partial_h \L)^{\top}.
\end{equation}
Actually, this is always true when $C=2$ and $w_k$'s are orthogonal.
When $C>2$, the approximate error is of order $\frac{e^{E_{i\ne t}}}{\sum_k e^{E_k}}$ and will be approaching zero during the training.
The remaining discussion is the same as the MSE case after Eq. \eqref{eq:mse1ststep}.

\section{The NegNet}
\label{sec:negnet}

During the search for DC decompositions of the loss, we encountered the following ``quasi-DC decomposition"
\begin{equation}
    G_3=\L'_{1/2}+\frac{\rho}{2} L_2,\;H_3=\L'_{1/2}-\L+\frac{\rho}{2} L_2,
\end{equation}
while ``quasi" means that only $G_3$ is convex but not $H_3$.
However, this decomposition could also explain the ResNet gradient.
Sometimes, quasi-DC decomposition works even better than DC decomposition; only its convergence is not guaranteed in theory.
Akin to the discussion around Eq. \eqref{eq:mse2ndstep}, we need to solve the following convex subproblem
\begin{equation}
    \argmin{\partial_w \L\, \Delta w+\frac{1}{2}\Delta w^{\top} \partial_w^2 G_3 \Delta w}
\end{equation}
where the Hessian of $G_3$ is almost of rank 1
\begin{equation}
    \frac{1}{2}\partial_w^2 G_3 -\frac{\rho}{4}=\frac{1}{4}(\ba+\bc)(\ba+\bc)^{\top}.
\end{equation}
So when we do the quasi-DCA iteration, we shall go along the $\ba+\bc$ direction.
This strategy means that by ignoring the higher-order terms (cubic terms) in Newton's method and considering them as noise compared to the second-order terms, the DC philosophy advocates for updates in the direction where this noise is minimized.
Because other directions possess $O(\Delta w^3)$ fluctuations, the DC philosophy now is to ``go along the direction with the least noise."

\begin{figure}[ht]
    \centering
    \includegraphics[width=1.0\linewidth]{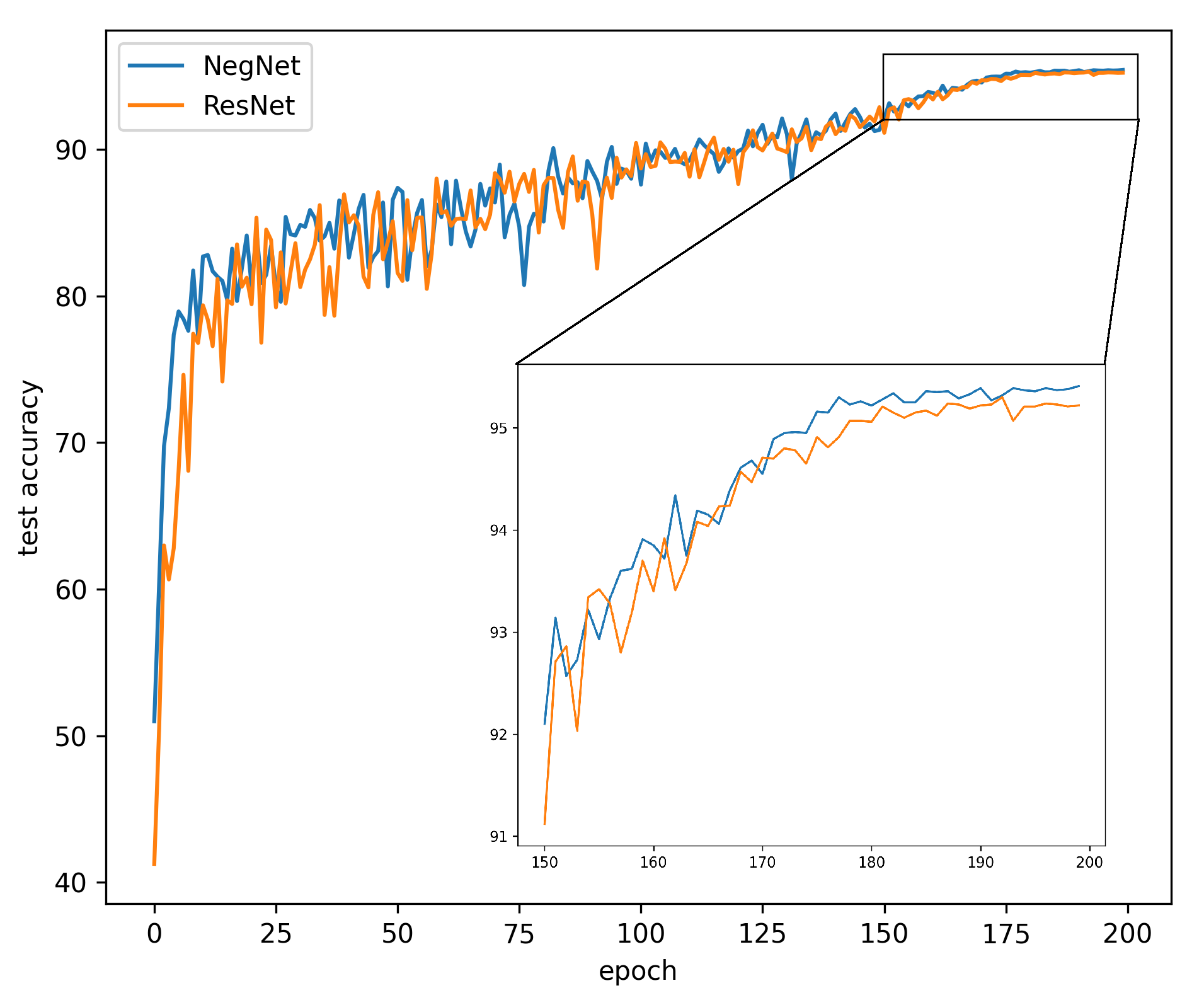}
    \caption{Performance comparison between NegNet18 (Eq. \eqref{eq:negnet}) and ResNet18 on the CIFAR10 dataset. The results show that NegNet achieves performance on par with ResNet. Note that the activation layer has been moved before the shortcut in NegNet for consistency with Eq. \eqref{eq:resnet}.}
    \label{fig:negnet}
\end{figure}

And if this quasi-DC decomposition works, the following quasi-DC decomposition should also work
\begin{equation}
    G_4=\L'_{-1/2}+\frac{\rho}{2} L_2,\;H_4=\L'_{-1/2}-\L+\frac{\rho}{2} L_2.
\end{equation}
This decomposition corresponds to the following paradigm
\begin{equation}\label{eq:negnet}\begin{cases}
    h^1=-x^0+F^0(x^0),~\cdots&\\
    h^l=-h^{l-1}+F^{l-1}(h^{l-1}),~\cdots&\\
    h^L=-h^{L-1}+F^{L-1}(h^{L-1}),&\\
    \L=\loss(h^L,\target).
\end{cases}
\end{equation}
We call it NegNet (Negative ResNet).
The NegNet is ``not even wrong" from the philosophy of ``Residual."
However, from the perspective of quasi-DCA, it would work as well as the ResNet.
So we did the experiment shown in Fig. \ref{fig:negnet}, and the result turned out to support the DCA viewpoint.

This simple example highlights the potential of DCA in inspiring the design of new neural network architectures. We suggest that, following the philosophy of DCA, greater attention should also be given to 2nd-order derivatives when developing novel architectures.

\section{Related Work}
\label{sec:relatedwork}

Extensive practical applications have validated the effectiveness of shortcuts, yet little theoretical work explains this phenomenon.
The year after ResNet was invented, \cite{resnet2016} pointed out that shortcuts in residual networks facilitate the flow of information during both the forward and backward processes, which makes training more accessible and improves generalization.
Considering the flow of signals (zeroth-order derivatives) and gradients (first-order derivatives) has become a standard step in the design of neural network architectures.
\cite{ensembles2016} interprets ResNet as an ensemble of many paths of differing lengths.
\cite{meanfield2017} studied ResNet using mean field theory.
\cite{corr2018} studied the correlation between gradients. They showed that the gradients in architectures with shortcuts are far more resistant to shattering and decaying sublinearly.
\cite{failmodes2018} proved that ResNet architecture can avoid some failure modes as long as the initializing is done right.
\cite{controltheory2018} related ResNet to an optimal control problem and provided stability results.
\cite{tao2020} studies the influence of residual connections on the hypothesis complexity of neural networks for the covering number of their hypothesis space.

Morden CNN architectures, such as YOLOv9 \cite{yolov9}, use auxiliary branches to generate reliable gradients.
This is in line with the spirit of this article, which involves deriving the conclusion that gradients need improvement and then modifying the network architecture to obtain the required gradients using automatic differentiation tools.

\section{Conclusion}
\label{sec:conclusion}

In this paper, we approached neural network design from the perspective of the DCA framework. By applying DCA to a vanilla network, we demonstrated that, with a straightforward choice of decomposition, the DCA approach can reproduce the shortcut technique commonly used in neural networks for both MSE and CE loss functions.
Essentially, our findings reveal that applying DCA to a standard network replicates the gradient effects observed in shortcut networks. This discovery not only provides a fresh perspective on the role and functionality of shortcuts but also underscores the critical importance of analyzing second-order derivatives when designing neural network architectures.
Additionally, our exploration of ``NegNet", which challenges the conventional interpretation of shortcuts, exemplifies the practical application of DCA in network design. We propose that DCA offers a powerful framework for developing new architectures by systematically applying it to existing designs. Leveraging the theoretical foundations of DCA holds significant promise for advancing global convergence in neural network training.


\section*{Acknowledgements}

This work was supported by the Natural Science Foundation of China (11601327). YR and YH profoundly thank Babak Haghighat for his invaluable guidance and selfless support.
YR would thank Jiaxuan Guo for his insightful opinions.



\section*{Impact Statement}

This paper presents work whose goal is to advance the field of 
Machine Learning. There are many potential societal consequences 
of our work, none of which we feel must be specifically highlighted here.


\bibliography{main}
\bibliographystyle{icml2024}

\newpage
\appendix
\onecolumn


\end{document}